\documentclass[
hf,
]{ceurart}

\usepackage{booktabs} %
\usepackage{float}
\restylefloat{table}
\usepackage{enumitem}
\usepackage{multirow}
\usepackage{colortbl}

\usepackage[most]{tcolorbox}
\def\checkblue{\textcolor{blue}} %

\newcommand{\anatomy}[1]{\textcolor{magenta}{#1}}

\newcommand{\code}[1]{{\small\ttfamily#1}}

\definecolor{codegreen}{rgb}{0,0.6,0}
\definecolor{codegray}{rgb}{0.5,0.5,0.5}

\definecolor{backcolour}{RGB}{245,248,250}
\definecolor{emph}{RGB}{166,88,53}
\definecolor{nightblue}{RGB}{9,49,105}
\definecolor{keywords}{RGB}{207,33,46}
\definecolor{lightpurple}{RGB}{130,81,223}

\lstdefinestyle{mystyle}{
    backgroundcolor=\color{backcolour},   
    commentstyle=\color{codegreen},
    keywordstyle=\color{keywords},
    stringstyle=\color{nightblue},
    basicstyle=\fontsize{7}{8}\ttfamily,
    breakatwhitespace=true,         
    breaklines=true,                 
    captionpos=b,                    
    keepspaces=true,                 
    numberstyle=\tiny\color{codegray},
    numbersep=2pt,                  
    showspaces=false,                
    showstringspaces=false,
    showtabs=false,                  
    tabsize=2,
    moredelim=**[is][\color{red}]{@}{@},
    moredelim=**[is][\ttfamily\color{black}]{+}{+},
    emph={dsp,Example,sample,annotate,knn,crossval,generate,retrieve,retrieve\_ensemble,majority,fused_retrieval,Template, Transformation,rank,branch},
    emphstyle={\color{lightpurple}},
    linewidth=0.98\columnwidth,
    frame=tb,    
    xrightmargin=0pt,
    xleftmargin=0.23cm,
    numbers=left,
    aboveskip=0.4cm,
    belowskip=0.4cm,
}

\begin{document}

\conference{Workshop on Causal Neuro-symbolic Artificial Intelligence, June 01--5, 2025, Portoroz, Slovenia}

\title{Can LLMs Leverage Observational Data? Towards Data-Driven Causal Discovery with LLMs}

\author[1]{Yuni Susanti}[%
email=susanti.yuni@fujitsu.com,
]
\cormark[1]
\address[1]{Fujitsu Limited, Japan}

\author[2]{Michael Färber}[%
email=michael.faerber@tu-dresden.de
]
\address[2]{ScaDS.AI \& TU Dresden, Germany}

\begin{abstract}
Causal discovery traditionally relies on statistical methods applied to observational data, often requiring large datasets and assumptions about underlying causal structures. Recent advancements in Large Language Models (LLMs) have introduced new possibilities for causal discovery by providing domain expert knowledge. However, it remains unclear whether LLMs can effectively process observational data for causal discovery. In this work, we explore the potential of LLMs for data-driven causal discovery by integrating observational data for LLM-based reasoning. Specifically, we examine whether LLMs can effectively utilize observational data through two prompting strategies: \textit{pairwise} prompting and \textit{breadth first search} (BFS)-based prompting. In both approaches, we incorporate the observational data directly into the prompt to assess LLMs' ability to infer causal relationships from such data. Experiments on benchmark datasets show that incorporating observational data enhances causal discovery, boosting F1 scores by up to 0.11 point using both pairwise and BFS LLM-based prompting, while outperforming traditional statistical causal discovery baseline by up to 0.52 points. Our findings highlight the potential and limitations of LLMs for data-driven causal discovery, demonstrating their ability to move beyond textual metadata and effectively utilize observational data for more informed causal reasoning. Our studies lays the groundwork for future advancements toward fully LLM-driven causal discovery.

\end{abstract}

\maketitle
\vspace{-0.2cm}
\section{Introduction}
\label{sec:intro}
\vspace{-0.2cm}
Understanding cause-and-effect relationships is fundamental to scientific discovery and decision-making across various fields such as biomedical research, economics, and social sciences. Traditionally, causal discovery relies on statistical methods applied to observational data, often requiring large datasets and strong assumptions about causal structures. Despite such limitations, statistical-based methods such as constraint-based approaches (e.g., PC algorithms~\cite{PC10.7551/mitpress/1754.001.0001}) and score-based methods (e.g., GES~\cite{GESChickering02a}), are still widely used in causal discovery. 

Recent advances in Large Language Models (LLMs) have opened new possibilities for causal discovery. LLMs have been primarily used as expert in \textit{knowledge-based causal discovery}, leveraging metadata—such as variable names and textual descriptions—to infer causal relationships~\cite{kic2023causal, SusantiKGSP}. However, this approach is limited by the quality and specificity of metadata, and the internal knowledge of the LLMs themselves making it prone to inconsistencies and domain-specific biases. With LLMs advancing in reasoning~\cite{wei2022emergentabilitieslargelanguage,wei2023chainofthoughtpromptingelicitsreasoning,NEURIPS2022_8bb0d291}, especially in text-based inference, a natural question emerges:
\begin{quote}
    \textbf{\textit{Can LLMs leverage observational data for causal discovery?}}
\end{quote}

Despite the importance of observational data in statistical causal discovery, existing LLM-based methods have yet to fully utilize it.
To address this gap, we propose a \textbf{data-driven causal discovery approach} that integrates observational data into LLM-based causal reasoning. We introduce prompting strategies incorporating observational data into the causal discovery process. By systematically embedding observational data into the prompts, we explore whether LLMs can enhance causal discovery beyond metadata-based inference, without relying solely on the LLMs' pre-existing domain knowledge or textual contexts. Our experiments across multiple benchmark datasets show that incorporating observational data improve LLMs' performance, up to 0.15 points in F1 scores, and outperform statistical-based methods. 
These results suggest that LLMs demonstrate potential in utilizing observational data for causal discovery, marking progress toward a hybrid model that integrates statistical methods with natural language reasoning via LLMs to better interpret data patterns for causal insights.

\vspace{-0.2cm}
\section{Related Work}
\label{relwork}
\vspace{-0.2cm}
LLMs have recently been utilized as expert systems for causal discovery, primarily by reasoning over \textit{metadata} of the variables rather than directly analyzing observational data. This approach, known as \textit{knowledge-based causal discovery}~\cite{kic2023causal, SusantiKGSP}, leverages LLMs’ ability to interpret domain-specific metadata—such as variable names and textual descriptions--to infer causal relationships. A widely adopted method in this paradigm is \textit{pairwise} prompting~\cite{susanti2024promptbasedvsfinetunedllms,kic2023causal}, where an LLM is systematically queried about the causal relationship between each pair of variables. This iterative process constructs a causal graph using LLM-derived insights, demonstrating promising results despite not incorporating observational data. 
Recent studies~\cite{kic2023causal,tu2023causaldiscovery,willig2022foundation,zhang2023understanding,gao-etal-2023-chatgpt} show that LLMs effectively provide background knowledge for causal discovery and outperform traditional non-LLM approaches.
Other research has evaluated LLMs' ability to identify causal relationships in text~\cite{khetan-etal-2022-mimicause,SusantiCEG,chatwal-etal-2025-enhancing}. For instance, recent work by~\cite{SusantiKGSP} introduced a method that integrates knowledge graph structures into LLM prompts to enhance causal relation extraction by smaller models.

A different line of research integrates LLMs with traditional causal discovery methods~\cite{takayama2024integratinglargelanguagemodels, ban2023querytoolscausalarchitects, 2024iclrcausal}. These approaches typically use LLMs to extract prior knowledge or serve as feedback agents to refine causal graphs. Some studies further examine how observational data can be used to improve LLMs' causal reasoning, such as by incorporating statistics calculated from observational data like Pearson correlation into the prompt~\cite{jiralerspong2024efficientcausalgraphdiscovery}. Unlike previous work, our work focuses on leveraging observational data \textit{directly} for LLM-based causal discovery. Rather than using LLMs solely as knowledge extraction tools or supplementary components for traditional methods, we investigate their ability to infer causal relationships by reasoning directly over structured observational data. This approach aims to push the boundaries of LLMs for data-driven causal discovery, demonstrating their potential as standalone reasoning agents.

\vspace{-0.2cm}
\section{Approach}
\label{sec:approach}
\vspace{-0.2cm}
\subsection{Task Formulation}
\label{sec:task}
\vspace{-0.2cm}
Given a set of observed variables \( \mathcal{V} = \{V_1, V_2, \dots, V_n\} \), the objective is to infer a \textbf{causal graph} \( \mathcal{G} = (\mathcal{V}, \mathcal{E}) \), where \( \mathcal{E} \subseteq \mathcal{V} \times \mathcal{V} \) represents directed causal relationships between variables, and \( V_i \) represents a node in the causal graph. We formulate our task as a \textit{classification} task where each pair of variables \( (V_i, V_j) \) must be classified as (1) \textit{$V_i$ causes $V_j$}, (2) \textit{$V_j$ causes $V_i$}, or (3) \textit{neither--no causal relationship}. The causal discovery process is then conducted using an LLM by formulating structured natural language prompts \( \mathcal{P} \) to elicit causal dependencies.

\vspace{-0.2cm}
\subsection{Data-Driven Causal Discovery with LLMs}
\label{sec:datadriven}
\vspace{-0.2cm}
Our approach to causal discovery with LLMs extends beyond existing knowledge-based method by integrating observational data \( \mathcal{D} \) into the prompt. However, since \( \mathcal{D} \) is often too large to fit within the prompt, we apply sampling function \( S \) to extract a representative subset \( \mathcal{D}_s \):

\begin{equation}
    \mathcal{D}_s = S(\mathcal{D}, k)
\end{equation}

where \( S(\cdot) \) is a sampling strategy (e.g., random, systematic, or cluster sampling), and \( k \) is the sample size constrained by the prompt length. The prompt \( \mathcal{P} \) then may encompass both prior knowledge \( \mathcal{K} \)--including known causal edges or constraints-- and the sampled data \( \mathcal{D}_s \). LLM's ability to infer causal relationships from such structured data distributions serves as the foundation of our data-driven causal discovery approach. In the following, we elaborate on the details of our proposed approach for systematically incorporating observational data into the prompt, utilizing (1) \textit{pairwise} and (2) \textit{BFS} prompting methods. 
\vspace{-0.3cm}
\paragraph{Pairwise Prompting with Observational Data.}
\label{sec:pairw} 
Pairwise Prompting is a localized approach where the LLM is queried about causal relationships between individual variable pairs. Given a variable pair \( (V_i, V_j) \), the LLM is instructed to determine whether a causal relationship exists between them, considering sampled observational data. The prompt $\mathcal{P}(V_i, V_j, \mathcal{K}, \mathcal{D}_s)$ in pairwise prompting explicitly asks:
\vspace{-0.2cm}
\begin{itemize}[leftmargin=0.5cm]
  \setlength\itemsep{-0.1em}
    \item \textbf{Existence:} Does \( V_i \) cause \( V_j \)?
    \item \textbf{Directionality:} If a causal relationship exists, is it \( V_i \to V_j \) or \( V_j \to V_i \)?
\end{itemize}
\vspace{-0.2cm}
The LLM then predicts the causal relationship by selecting from three options: \textit{\( V_i \) causes \( V_j \)}, \textit{\( V_j \) causes \( V_i \)}, or \textit{neither—no causal relationship}, as illustrated in Figure 1 (\textit{left}).

\vspace{-0.3cm}
\paragraph{BFS Prompting with Observational Data.}
\label{sec:bfs} 
The pairwise prompting requires a quadratic number of queries, making it impractical for large graphs. To address this, \cite{jiralerspong2024efficientcausalgraphdiscovery} introduce a framework using \textit{breadth-first search} (BFS) strategy, reducing the number of queries to a linear scale. Instead of analyzing pairs, the LLM explores causal relationship by traversing the graph using BFS technique.
In this work, we apply BFS prompting by~\cite{jiralerspong2024efficientcausalgraphdiscovery}, consisting three stages: 
\vspace{-0.2cm}
\begin{enumerate}[leftmargin=0.5cm]
\setlength\itemsep{-0.1em}
    \item \textbf{Initialization} – The LLM identifies variables that are not causally influenced by others. 
    \item \textbf{Expansion} – The LLM determines which variables are caused by the current node. 
    \item \textbf{Insertion} – The proposed variables are added to the BFS queue, and suggested edges are checked for cycles before being inserted. 
\end{enumerate}
\vspace{-0.2cm}
Figure 1 (\textit{right}) illustrates a BFS approach with observational data. Unlike the pairwise approach, the LLM directly responds with variables instead of selecting from given options.

\begin{figure}[ht]
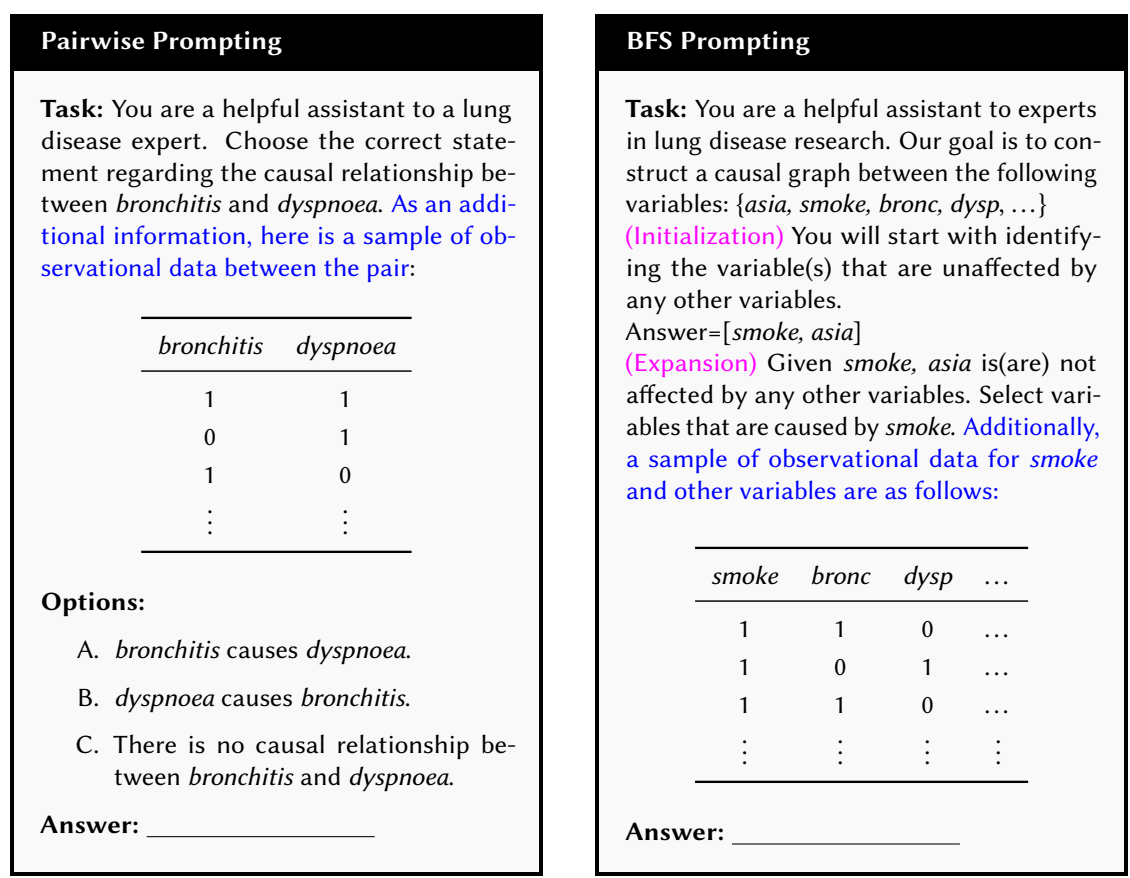

    \centering
    \begin{tcbraster}[raster columns=2,raster equal height,nobeforeafter,raster column skip=0.7cm]
    \begin{tcolorbox}[
    title=\textbf{Pairwise Prompting}, 
    colback=gray!5, 
    colframe=black, 
    fonttitle=\bfseries,
    boxsep=5pt, 
    left=5pt, 
    right=5pt, 
    top=5pt, 
    bottom=5pt,
    sharp corners ]
    \textbf{Task:} You are a helpful assistant to a lung disease expert. Choose the correct statement regarding the causal relationship between \textit{bronchitis} and \textit{dyspnoea}. \checkblue{As an additional information, here is a sample of observational data between the pair}:
    \vspace{3pt}
    \renewcommand{\arraystretch}{1.2}
    \begin{center}
        \begin{tabular}{c c}
            \toprule
            \textit{bronchitis} & \textit{dyspnoea} \\
            \midrule
            1 & 1 \\
            0 & 1 \\
            1 & 0 \\
            \vdots & \vdots \\
            \bottomrule
        \end{tabular}
    \end{center}
    \vspace{5pt}
    \textbf{Options:}
    \begin{enumerate}
        \item[A.] \textit{bronchitis} causes \textit{dyspnoea}.
        \item[B.] \textit{dyspnoea} causes \textit{bronchitis}.
        \item[C.] There is no causal relationship between \textit{bronchitis} and \textit{dyspnoea}.
    \end{enumerate}
    \textbf{Answer:} \underline{\hspace{3cm}}
    \end{tcolorbox}
      \begin{tcolorbox}[
    title=\textbf{BFS Prompting}, 
    colback=gray!5, 
    colframe=black, 
    fonttitle=\bfseries,
    boxsep=5pt, 
    left=5pt, 
    right=5pt, 
    top=5pt, 
    bottom=5pt,
    sharp corners ]
    \textbf{Task:} You are a helpful assistant to experts in lung disease research. Our goal is to construct a causal graph between the following variables: \{\textit{asia, smoke, bronc, dysp}, \ldots\} \\
    \anatomy{(Initialization)} You will start with identifying the variable(s) that are unaffected by any other variables. \\
    Answer=[\textit{smoke, asia}] \\
    \anatomy{(Expansion)} Given \textit{smoke, asia} is(are) not affected by any other variables. Select variables that are caused by \textit{smoke}. \checkblue{Additionally, a sample of observational data for \textit{smoke} and other variables are as follows:}
    \vspace{5pt}
    \vspace{3pt}
    \renewcommand{\arraystretch}{1.2}
    \begin{center}
        \begin{tabular}{c c c c}
            \toprule
            \textit{smoke} & \textit{bronc} & \textit{dysp} & \ldots \\
            \midrule
            1 & 1 & 0 & \ldots \\
            1 & 0 & 1 & \ldots \\
            1 & 1 & 0 & \ldots \\
            \vdots & \vdots & \vdots & \vdots \\
            \bottomrule
        \end{tabular}
    \end{center}
    \vspace{5pt}
    \textbf{Answer:} \underline{\hspace{3cm}}
    \end{tcolorbox}
    \end{tcbraster}
    \label{fig:prompt}
    \caption{Prompt examples for Pairwise and BFS prompting~\cite{jiralerspong2024efficientcausalgraphdiscovery} using observational data.}
\end{figure}

\vspace{-0.2cm}
\section{Evaluation}
\vspace{-0.2cm}
\subsection{Evaluation Settings}
\label{sect:data}
\vspace{-0.2cm}
\paragraph{Dataset.} We conduct experiments on datasets from \code{BNLearn}~\cite{Scutari2010}, a collection of Bayesian network datasets widely used for testing causal discovery algorithms, as follows:
\vspace{-0.2cm}
\begin{enumerate}[leftmargin=0.5cm]
\setlength\itemsep{-0.1em}
    \item \textsf{\textbf{ASIA~\cite{asia}}}: A network with 8 variables (e.g., \textit{dyspnoea}, \textit{bronchitis}, and \textit{if a patient has recently traveled to Asia}), for lung diseases diagnosis based on medical observations. 
    \item \textsf{\textbf{CANCER~\cite{Scutari2010}}}: A network that models the factors influencing cancer development. It contains fewer variables than \textsf{ASIA}, but with more intricate dependency structures.
    \item \textsf{\textbf{SURVEY~\cite{Scutari2010}}}: A dataset on how public transport usage varies across social groups, based on survey responses, with variables such as \textit{age}, \textit{occupation}, and \textit{preferred means of transportation}. 
\end{enumerate}
\vspace{-0.2cm}
Despite their modest size, we specifically selected them because they offer valuable insights for evaluating causal discovery in straightforward, well-defined relationships.
\vspace{-0.2cm}
\paragraph{Model Comparison.} We compare LLM-based causal discovery against statistical causal discovery methods, including: 
\textbf{\textsf{(1) PC Algorithm}}~\cite{PC10.7551/mitpress/1754.001.0001} and 
\textbf{\textsf{(2) GES}}~\cite{GESChickering02a}.
For LLM-based causal discovery, we compare the two prompting strategies, with variations that include and exclude observational data as an additional input:  
\textbf{\textsf{(3) Pairwise Prompting, (4) Pairwise Prompting + Observational Data, (5) BFS Prompting, (6) BFS Prompting + Observational Data}}.
Additionally, we incorporate \textit{pearson correlation} calculated from the observational data, following \cite{jiralerspong2024efficientcausalgraphdiscovery}:  \textbf{\textsf{(7) Pairwise Prompting + Pearson corr., (8) BFS Prompting + Pearson corr.}}.

\vspace{-0.2cm}
\paragraph{Experimental Setup.} For each dataset, we conduct experiments across varying sample sizes in $\{100, 500, 1000\}$
for statistical-based methods, while keeping LLM-based methods fixed at $k$=100 observational data due to token length limitations. The samples were selected using various sampling strategy $S$ —\textit{random}, \textit{cluster}, \textit{systematic}, and \textit{adaptive (K-means)} sampling methods.
However, since the results showed no significant differences, we reported the scores from \textit{simple random} sampling.
We used \texttt{GPT} model \code{gpt-4-0125-preview} checkpoint, query it four times varying sampling temperatures in $\{0, 0.5, 0.7, 1.0\}$ and report the average results. We adapted the implementation code from the original BFS prompting paper~\cite{jiralerspong2024efficientcausalgraphdiscovery} for our experiment, which includes implementation of PC~\cite{PC10.7551/mitpress/1754.001.0001} and GES~\cite{GESChickering02a} from \code{causal-learn} package~\cite{zheng2024causal}.

\begin{table*}
    \centering
    \small
    \tabcolsep=0.15cm
    \begin{tabular}{lccc|ccc|ccc}
    \toprule
        & \multicolumn{3}{c|}{\textsf{ASIA}} & \multicolumn{3}{c|}{\textsf{CANCER}} & \multicolumn{3}{c}{\textsf{SURVEY}} \\
        \cline{2-10}
        & F1$\uparrow$ & NHD$\downarrow$ & Ratio$\downarrow$ & F1$\uparrow$ & NHD$\downarrow$ & Ratio$\downarrow$ & F1$\uparrow$ & NHD$\downarrow$ & Ratio$\downarrow$ \\
        \midrule
        \multicolumn{1}{l}{\textbf{Statistical-based Methods}} & \multicolumn{9}{c}{} \\
        \midrule
        PC~\cite{PC10.7551/mitpress/1754.001.0001} & 0.50 & 0.24 & 0.50 & 0.33 & 0.44 & 0.67 & 0.50 & 0.44 & 0.50 \\
        GES~\cite{GESChickering02a} & 0.38 & 0.28 & 0.63 & 0.33 & 0.44 & 0.67 & 0.25 & 0.10 & 0.75 \\
        \midrule
        \multicolumn{1}{l}{\textbf{LLM-based Methods}} & \multicolumn{9}{c}{} \\
        \midrule
        Pairwise Prompting & 0.47 & 0.29 & 0.53 & 0.60 & 0.22 & 0.39 & 0.45 & 0.31 & 0.55 \\
        +Pearson corr. & 0.64 & 0.13 & \underline{0.36} & \underline{0.67} & 0.16 & \underline{0.33} & 0.20 & 0.32 & 0.80 \\
        \rowcolor[HTML]{ECF4FF}+Observational Data & \underline{0.58} & 0.16 & 0.42 & 0.66 & 0.18 & 0.35 & \underline{0.53} & 0.19 & \underline{0.47} \\
        \midrule
        BFS Prompting~\cite{jiralerspong2024efficientcausalgraphdiscovery} & 0.85 & 0.04 & 0.15 & 0.66 & 0.12 & 0.33 & 0.50 & 0.22 & 0.50 \\
        +Pearson corr. & 0.88 & 0.03 & 0.12 & 0.72 & 0.12 & 0.27 & 0.45 & 0.31 & 0.55 \\
        \rowcolor[HTML]{ECF4FF}+Observational Data & \anatomy{\underline{0.90}} & 0.03 & \anatomy{\underline{0.10}} & \anatomy{\underline{0.77}} & 0.10 & \anatomy{\underline{0.23}} & \anatomy{\underline{0.54}} & 0.20 & \anatomy{\underline{ 0.45}} \\
        \bottomrule
    \end{tabular}
    \caption{Performance comparison on benchmark datasets. The best scores are marked in \anatomy{pink}. For LLM-based approaches, we queried the model four times and reported the \textbf{average} scores.}
    \label{tab:results}
\end{table*}
    
\vspace{-0.2cm}
\subsection{Results and Discussion}
\label{sec:resultdis}
\vspace{-0.2cm}
Table~\ref{tab:results} summarizes our experiment results. Since we frame causal discovery as a classification task, we compute classification metrics e.g., Precision, Recall, F1 score and report \textit{normalized hamming distance} (NHD) and \textit{ratio}, following~\cite{kic2023causal,jiralerspong2024efficientcausalgraphdiscovery}. We discuss the key findings as follows:

\vspace{-0.2cm}
\paragraph{\textbf{LLM-based methods outperform statistical-based methods in most cases.}}
Across all datasets, LLM-based methods including both Pairwise and BFS-based prompting show significant improvements over PC and GES in terms of F1 score. The improvement is especially significant in BFS-based prompting with observational data, achieving a 0.44-point increase (0.33 vs. 0.77 on \textsf{CANCER}) compared to PC method. Similarly, on \textsf{ASIA}, it delivers a 0.40-point gain (0.50 vs.0.90). When comparing LLM-based methods to GES, we observe a consistent F1 score improvement ranging from 0.29 to 0.44 across all datasets, highlighting the effectiveness of our approach.

While PC and GES are well-established for causal discovery, their performance heavily depends on sample size. In our experiments, we set a fixed number of 100 samples for LLM-based methods across all datasets, whereas statistical-based methods (PC and GES) are evaluated with varying sample sizes $\{100, 500, 1000\}$. Our results show that LLM-based methods, particularly when enriched with observational data, achieve strong performance even with a limited number of samples. In contrast, statistical-based methods may require as many as 1000 samples to reach comparable performance, as observed in the \textsf{SURVEY} dataset, where the PC method matches LLM-based methods at 1000 sample size. This underscores the robustness of LLM-based causal discovery, making it particularly valuable in data-limited scenarios.

\vspace{-0.2cm}
\paragraph{\textbf{Observational data improves LLM-based methods' performance.}}
Across both LLM prompting methods, incorporating observational data consistently enhances F1 scores while reducing NHD and Ratio values, demonstrating its effectiveness in improving causal discovery. In Pairwise Prompting, adding observational data results in a F1 score increase of up to 0.11 points (0.47 to 0.58 on \textsf{ASIA}). Similarly, in BFS-based Prompting, adding observational data leads to the best overall performance across all datasets, with F1 scores improving by up to 0.11 points (0.66 to 0.77 on \textsf{CANCER}). These results suggest that, despite being primarily trained using text-based data, LLMs demonstrate a potential to effectively leverage observational numerical data as contextual grounding for causal discovery and reasoning. 

Additionally, we assess the impact of incorporating Pearson correlation derived from the same observational data, following \cite{jiralerspong2024efficientcausalgraphdiscovery}. The results demonstrate a consistent improvement over methods without any additional information. However, we find that our method of directly adding observational data yields better overall performance on average (0.075 vs. 0.035). This suggests that by leveraging numerical insights, LLMs make more informed causal inferences and providing observational data reduces their reliance on surface-level textual patterns, further bridging the gap between data-driven and knowledge-driven approaches in causal discovery.
\vspace{-0.2cm}
\paragraph{\textbf{BFS Prompting consistently outperforms Pairwise Prompting.}}
BFS Prompting achieves the highest F1 scores and lowest Ratio values in all datasets (values marked in \anatomy{pink} in Table~\ref{tab:results}), demonstrating its superior ability to leverage observational data for causal discovery. 
Beyond being more efficient than its pairwise counterpart, BFS-based prompting demonstrates its superiority (up to 0.32 point, $0.58 vs. 0.90$ on \textsf{ASIA}) by offering a more contextual and structured approach to causal discovery. This suggests that leveraging global context awareness—multi-variable interactions rather than variable pairs in isolation—enhances causal inference. However, this prompting method includes the entire query history, which can lead to excessive prompt length and may be infeasible due to the LLM’s token limitations.

\vspace{-0.2cm}
\section{Conclusion}
\label{sec:conclusion}
\vspace{-0.2cm}
In this work, we investigated the potential of Large Language Models (LLMs) for data-driven causal discovery by integrating observational data into their reasoning process. Through experiments on causal benchmark datasets, we assessed the extent to which LLMs can infer causal relationships from structured, observational data. Our results suggest that LLMs demonstrate potential in utilizing observational data for causal discovery, marking progress toward a hybrid model that integrates statistical methods with natural language reasoning with LLMs.

Despite these promising results, the effectiveness of LLMs is still dataset-dependent, and reasoning stability can vary. 
Future work should further explore this hybrid approaches of LLM-based reasoning with statistical causal discovery, as well as refining prompting strategies and sampling selection approach. Additionally, it would be valuable to investigate performance across multiple LLMs and extending on larger dataset.
By continuously improving LLMs' ability to process structured data, we move toward a more comprehensive framework that unifies statistical causal discovery with the reasoning capabilities of LLMs.

\bibliography{ref}

\begin{thebibliography}{22}
\expandafter\ifx\csname natexlab\endcsname\relax\def\natexlab#1{#1}\fi
\providecommand{\url}[1]{\texttt{#1}}
\providecommand{\href}[2]{#2}
\providecommand{\path}[1]{#1}
\providecommand{\DOIprefix}{doi:}
\providecommand{\ArXivprefix}{arXiv:}
\providecommand{\URLprefix}{URL: }
\providecommand{\Pubmedprefix}{pmid:}
\providecommand{\doi}[1]{\href{http://dx.doi.org/#1}{\path{#1}}}
\providecommand{\Pubmed}[1]{\href{pmid:#1}{\path{#1}}}
\providecommand{\bibinfo}[2]{#2}
\ifx\xfnm\relax \def\xfnm[#1]{\unskip,\space#1}\fi
\bibitem[{Spirtes et~al.(2001)Spirtes, Glymour, and Scheines}]{PC10.7551/mitpress/1754.001.0001}
\bibinfo{author}{P.~Spirtes}, \bibinfo{author}{C.~Glymour}, \bibinfo{author}{R.~Scheines}, \bibinfo{title}{Causation, Prediction, and Search}, \bibinfo{publisher}{The MIT Press}, \bibinfo{year}{2001}. \URLprefix \url{https://doi.org/10.7551/mitpress/1754.001.0001}.
\bibitem[{Chickering(2002)}]{GESChickering02a}
\bibinfo{author}{D.~M. Chickering},
\newblock \bibinfo{title}{Optimal structure identification with greedy search.},
\newblock \bibinfo{journal}{J. Mach. Learn. Res.} \bibinfo{volume}{3} (\bibinfo{year}{2002}) \bibinfo{pages}{507--554}. \URLprefix \url{http://dblp.uni-trier.de/db/journals/jmlr/jmlr3.html#Chickering02a}.
\bibitem[{Kıcıman et~al.(2023)Kıcıman, Ness, Sharma, and Tan}]{kic2023causal}
\bibinfo{author}{E.~Kıcıman}, \bibinfo{author}{R.~Ness}, \bibinfo{author}{A.~Sharma}, \bibinfo{author}{C.~Tan}, \bibinfo{title}{Causal reasoning and large language models: Opening a new frontier for causality}, \bibinfo{year}{2023}.
\bibitem[{Susanti and F{\"a}rber(2025)}]{SusantiKGSP}
\bibinfo{author}{Y.~Susanti}, \bibinfo{author}{M.~F{\"a}rber},
\newblock \bibinfo{title}{Knowledge graph structure as prompt: Improving small language models capabilities for knowledge-based causal discovery},
\newblock in: \bibinfo{editor}{G.~Demartini}, \bibinfo{editor}{K.~Hose}, \bibinfo{editor}{M.~Acosta}, \bibinfo{editor}{M.~Palmonari}, \bibinfo{editor}{G.~Cheng}, \bibinfo{editor}{H.~Skaf-Molli}, \bibinfo{editor}{N.~Ferranti}, \bibinfo{editor}{D.~Hern{\'a}ndez}, \bibinfo{editor}{A.~Hogan} (Eds.), \bibinfo{booktitle}{The Semantic Web -- ISWC 2024}, \bibinfo{publisher}{Springer Nature Switzerland}, \bibinfo{address}{Cham}, \bibinfo{year}{2025}, pp. \bibinfo{pages}{87--106}.
\bibitem[{Wei et~al.(2022)Wei, Tay, Bommasani, Raffel, Zoph, Borgeaud, Yogatama, Bosma, Zhou, Metzler, Chi, Hashimoto, Vinyals, Liang, Dean, and Fedus}]{wei2022emergentabilitieslargelanguage}
\bibinfo{author}{J.~Wei}, \bibinfo{author}{Y.~Tay}, \bibinfo{author}{R.~Bommasani}, \bibinfo{author}{C.~Raffel}, \bibinfo{author}{B.~Zoph}, \bibinfo{author}{S.~Borgeaud}, \bibinfo{author}{D.~Yogatama}, \bibinfo{author}{M.~Bosma}, \bibinfo{author}{D.~Zhou}, \bibinfo{author}{D.~Metzler}, \bibinfo{author}{E.~H. Chi}, \bibinfo{author}{T.~Hashimoto}, \bibinfo{author}{O.~Vinyals}, \bibinfo{author}{P.~Liang}, \bibinfo{author}{J.~Dean}, \bibinfo{author}{W.~Fedus}, \bibinfo{title}{Emergent abilities of large language models}, \bibinfo{year}{2022}. \URLprefix \url{https://arxiv.org/abs/2206.07682}. \href{http://arxiv.org/abs/2206.07682}{{\tt arXiv:2206.07682}}.
\bibitem[{Wei et~al.(2023)Wei, Wang, Schuurmans, Bosma, Ichter, Xia, Chi, Le, and Zhou}]{wei2023chainofthoughtpromptingelicitsreasoning}
\bibinfo{author}{J.~Wei}, \bibinfo{author}{X.~Wang}, \bibinfo{author}{D.~Schuurmans}, \bibinfo{author}{M.~Bosma}, \bibinfo{author}{B.~Ichter}, \bibinfo{author}{F.~Xia}, \bibinfo{author}{E.~Chi}, \bibinfo{author}{Q.~Le}, \bibinfo{author}{D.~Zhou}, \bibinfo{title}{Chain-of-thought prompting elicits reasoning in large language models}, \bibinfo{year}{2023}. \URLprefix \url{https://arxiv.org/abs/2201.11903}. \href{http://arxiv.org/abs/2201.11903}{{\tt arXiv:2201.11903}}.
\bibitem[{Kojima et~al.(2022)Kojima, Gu, Reid, Matsuo, and Iwasawa}]{NEURIPS2022_8bb0d291}
\bibinfo{author}{T.~Kojima}, \bibinfo{author}{S.~S. Gu}, \bibinfo{author}{M.~Reid}, \bibinfo{author}{Y.~Matsuo}, \bibinfo{author}{Y.~Iwasawa},
\newblock \bibinfo{title}{Large language models are zero-shot reasoners},
\newblock in: \bibinfo{editor}{S.~Koyejo}, \bibinfo{editor}{S.~Mohamed}, \bibinfo{editor}{A.~Agarwal}, \bibinfo{editor}{D.~Belgrave}, \bibinfo{editor}{K.~Cho}, \bibinfo{editor}{A.~Oh} (Eds.), \bibinfo{booktitle}{Advances in Neural Information Processing Systems}, volume~\bibinfo{volume}{35}, \bibinfo{publisher}{Curran Associates, Inc.}, \bibinfo{year}{2022}, pp. \bibinfo{pages}{22199--22213}. \URLprefix \url{https://proceedings.neurips.cc/paper_files/paper/2022/file/8bb0d291acd4acf06ef112099c16f326-Paper-Conference.pdf}.
\bibitem[{Susanti and Holsmoelle(2025)}]{susanti2024promptbasedvsfinetunedllms}
\bibinfo{author}{Y.~Susanti}, \bibinfo{author}{N.~Holsmoelle}, \bibinfo{title}{Prompting or fine-tuning? exploring large language models for causal graph validation}, \bibinfo{year}{2025}. \URLprefix \url{https://arxiv.org/abs/2406.16899}. \href{http://arxiv.org/abs/2406.16899}{{\tt arXiv:2406.16899}}.
\bibitem[{Tu et~al.(2023)Tu, Ma, and Zhang}]{tu2023causaldiscovery}
\bibinfo{author}{R.~Tu}, \bibinfo{author}{C.~Ma}, \bibinfo{author}{C.~Zhang}, \bibinfo{title}{Causal-discovery performance of chatgpt in the context of neuropathic pain diagnosis}, \bibinfo{year}{2023}. \href{http://arxiv.org/abs/2301.13819}{{\tt arXiv:2301.13819}}.
\bibitem[{Willig et~al.(2022)Willig, Zečević, Dhami, and Kersting}]{willig2022foundation}
\bibinfo{author}{M.~Willig}, \bibinfo{author}{M.~Zečević}, \bibinfo{author}{D.~S. Dhami}, \bibinfo{author}{K.~Kersting}, \bibinfo{title}{Can foundation models talk causality?}, \bibinfo{year}{2022}. \href{http://arxiv.org/abs/2206.10591}{{\tt arXiv:2206.10591}}.
\bibitem[{Zhang et~al.(2023)Zhang, Bauer, Bennett, Gao, Gong, Hilmkil, Jennings, Ma, Minka, Pawlowski, and Vaughan}]{zhang2023understanding}
\bibinfo{author}{C.~Zhang}, \bibinfo{author}{S.~Bauer}, \bibinfo{author}{P.~Bennett}, \bibinfo{author}{J.~Gao}, \bibinfo{author}{W.~Gong}, \bibinfo{author}{A.~Hilmkil}, \bibinfo{author}{J.~Jennings}, \bibinfo{author}{C.~Ma}, \bibinfo{author}{T.~Minka}, \bibinfo{author}{N.~Pawlowski}, \bibinfo{author}{J.~Vaughan}, \bibinfo{title}{Understanding causality with large language models: Feasibility and opportunities}, \bibinfo{year}{2023}. \href{http://arxiv.org/abs/2304.05524}{{\tt arXiv:2304.05524}}.
\bibitem[{Gao et~al.(2023)Gao, Ding, Qin, and Liu}]{gao-etal-2023-chatgpt}
\bibinfo{author}{J.~Gao}, \bibinfo{author}{X.~Ding}, \bibinfo{author}{B.~Qin}, \bibinfo{author}{T.~Liu},
\newblock \bibinfo{title}{Is {C}hat{GPT} a good causal reasoner? a comprehensive evaluation},
\newblock in: \bibinfo{editor}{H.~Bouamor}, \bibinfo{editor}{J.~Pino}, \bibinfo{editor}{K.~Bali} (Eds.), \bibinfo{booktitle}{Findings of the Association for Computational Linguistics: EMNLP 2023}, \bibinfo{publisher}{Association for Computational Linguistics}, \bibinfo{address}{Singapore}, \bibinfo{year}{2023}, pp. \bibinfo{pages}{11111--11126}. \URLprefix \url{https://aclanthology.org/2023.findings-emnlp.743}. \DOIprefix\doi{10.18653/v1/2023.findings-emnlp.743}.
\bibitem[{Khetan et~al.(2022)Khetan, Rizvi, Huber, Bartusiak, Sacaleanu, and Fano}]{khetan-etal-2022-mimicause}
\bibinfo{author}{V.~Khetan}, \bibinfo{author}{M.~I. Rizvi}, \bibinfo{author}{J.~Huber}, \bibinfo{author}{P.~Bartusiak}, \bibinfo{author}{B.~Sacaleanu}, \bibinfo{author}{A.~Fano},
\newblock \bibinfo{title}{{MIMIC}ause: {R}epresentation and automatic extraction of causal relation types from clinical notes},
\newblock in: \bibinfo{booktitle}{Findings of the Association for Computational Linguistics: ACL 2022}, \bibinfo{publisher}{Association for Computational Linguistics}, \bibinfo{address}{Dublin, Ireland}, \bibinfo{year}{2022}, pp. \bibinfo{pages}{764--773}. \URLprefix \url{https://aclanthology.org/2022.findings-acl.63}. \DOIprefix\doi{10.18653/v1/2022.findings-acl.63}.
\bibitem[{Susanti and Uchino(2024)}]{SusantiCEG}
\bibinfo{author}{Y.~Susanti}, \bibinfo{author}{K.~Uchino},
\newblock \bibinfo{title}{Causal-evidence graph for causal relation classification},
\newblock in: \bibinfo{booktitle}{Proceedings of the 39th ACM/SIGAPP Symposium on Applied Computing}, SAC '24, \bibinfo{publisher}{Association for Computing Machinery}, \bibinfo{address}{New York, NY, USA}, \bibinfo{year}{2024}, p. \bibinfo{pages}{714–722}. \URLprefix \url{https://doi.org/10.1145/3605098.3635894}. \DOIprefix\doi{10.1145/3605098.3635894}.
\bibitem[{Chatwal et~al.(2025)Chatwal, Agarwal, and Mittal}]{chatwal-etal-2025-enhancing}
\bibinfo{author}{P.~Chatwal}, \bibinfo{author}{A.~Agarwal}, \bibinfo{author}{A.~Mittal},
\newblock \bibinfo{title}{Enhancing causal relationship detection using prompt engineering and large language models},
\newblock in: \bibinfo{editor}{C.-C. Chen}, \bibinfo{editor}{A.~Moreno-Sandoval}, \bibinfo{editor}{J.~Huang}, \bibinfo{editor}{Q.~Xie}, \bibinfo{editor}{S.~Ananiadou}, \bibinfo{editor}{H.-H. Chen} (Eds.), \bibinfo{booktitle}{Proceedings of the Joint Workshop of the 9th Financial Technology and Natural Language Processing (FinNLP), the 6th Financial Narrative Processing (FNP), and the 1st Workshop on Large Language Models for Finance and Legal (LLMFinLegal)}, \bibinfo{publisher}{Association for Computational Linguistics}, \bibinfo{address}{Abu Dhabi, UAE}, \bibinfo{year}{2025}, pp. \bibinfo{pages}{248--252}. \URLprefix \url{https://aclanthology.org/2025.finnlp-1.26/}.
\bibitem[{Takayama et~al.(2024)Takayama, Okuda, Pham, Ikenoue, Fukuma, Shimizu, and Sannai}]{takayama2024integratinglargelanguagemodels}
\bibinfo{author}{M.~Takayama}, \bibinfo{author}{T.~Okuda}, \bibinfo{author}{T.~Pham}, \bibinfo{author}{T.~Ikenoue}, \bibinfo{author}{S.~Fukuma}, \bibinfo{author}{S.~Shimizu}, \bibinfo{author}{A.~Sannai}, \bibinfo{title}{Integrating large language models in causal discovery: A statistical causal approach}, \bibinfo{year}{2024}. \URLprefix \url{https://arxiv.org/abs/2402.01454}. \href{http://arxiv.org/abs/2402.01454}{{\tt arXiv:2402.01454}}.
\bibitem[{Ban et~al.(2023)Ban, Chen, Wang, and Chen}]{ban2023querytoolscausalarchitects}
\bibinfo{author}{T.~Ban}, \bibinfo{author}{L.~Chen}, \bibinfo{author}{X.~Wang}, \bibinfo{author}{H.~Chen}, \bibinfo{title}{From query tools to causal architects: Harnessing large language models for advanced causal discovery from data}, \bibinfo{year}{2023}. \URLprefix \url{https://arxiv.org/abs/2306.16902}. \href{http://arxiv.org/abs/2306.16902}{{\tt arXiv:2306.16902}}.
\bibitem[{Abdulaal et~al.(2024)Abdulaal, adamos hadjivasiliou, Montana-Brown, He, Ijishakin, Drobnjak, Castro, and Alexander}]{2024iclrcausal}
\bibinfo{author}{A.~Abdulaal}, \bibinfo{author}{adamos hadjivasiliou}, \bibinfo{author}{N.~Montana-Brown}, \bibinfo{author}{T.~He}, \bibinfo{author}{A.~Ijishakin}, \bibinfo{author}{I.~Drobnjak}, \bibinfo{author}{D.~C. Castro}, \bibinfo{author}{D.~C. Alexander}, \bibinfo{title}{Causal modelling agents: Causal graph discovery through synergising metadata- and data-driven reasoning}, \bibinfo{year}{2024}. \URLprefix \url{https://openreview.net/forum?id=pAoqRlTBtY}.
\bibitem[{Jiralerspong et~al.(2024)Jiralerspong, Chen, More, Shah, and Bengio}]{jiralerspong2024efficientcausalgraphdiscovery}
\bibinfo{author}{T.~Jiralerspong}, \bibinfo{author}{X.~Chen}, \bibinfo{author}{Y.~More}, \bibinfo{author}{V.~Shah}, \bibinfo{author}{Y.~Bengio}, \bibinfo{title}{Efficient causal graph discovery using large language models}, \bibinfo{year}{2024}. \URLprefix \url{https://arxiv.org/abs/2402.01207}. \href{http://arxiv.org/abs/2402.01207}{{\tt arXiv:2402.01207}}.
\bibitem[{Scutari(2010)}]{Scutari2010}
\bibinfo{author}{M.~Scutari},
\newblock \bibinfo{title}{Learning bayesian networks with the bnlearn r package},
\newblock \bibinfo{journal}{Journal of Statistical Software} \bibinfo{volume}{35} (\bibinfo{year}{2010}) \bibinfo{pages}{1--22}. \URLprefix \url{https://www.jstatsoft.org/v35/i03/}. \DOIprefix\doi{10.18637/jss.v035.i03}.
\bibitem[{Lauritzen and Spiegelhalter(2018)}]{asia}
\bibinfo{author}{S.~L. Lauritzen}, \bibinfo{author}{D.~J. Spiegelhalter},
\newblock \bibinfo{title}{Local computations with probabilities on graphical structures and their application to expert systems},
\newblock \bibinfo{journal}{Journal of the Royal Statistical Society: Series B (Methodological)} \bibinfo{volume}{50} (\bibinfo{year}{2018}) \bibinfo{pages}{157--194}. \DOIprefix\doi{10.1111/j.2517-6161.1988.tb01721.x}.
\bibitem[{Zheng et~al.(2024)Zheng, Huang, Chen, Ramsey, Gong, Cai, Shimizu, Spirtes, and Zhang}]{zheng2024causal}
\bibinfo{author}{Y.~Zheng}, \bibinfo{author}{B.~Huang}, \bibinfo{author}{W.~Chen}, \bibinfo{author}{J.~Ramsey}, \bibinfo{author}{M.~Gong}, \bibinfo{author}{R.~Cai}, \bibinfo{author}{S.~Shimizu}, \bibinfo{author}{P.~Spirtes}, \bibinfo{author}{K.~Zhang},
\newblock \bibinfo{title}{Causal-learn: Causal discovery in python},
\newblock \bibinfo{journal}{Journal of Machine Learning Research} \bibinfo{volume}{25} (\bibinfo{year}{2024}) \bibinfo{pages}{1--8}.

\end{thebibliography}

\end{document}